\documentclass[cameraready]{vgtc}

\ifpdf
  \pdfoutput=1\relax
  \usepackage{graphicx}
  \DeclareGraphicsExtensions{.pdf,.png,.jpg,.jpeg}
\else
  \usepackage{graphicx}
  \DeclareGraphicsExtensions{.eps}
\fi

\graphicspath{{figures/}{./}}

\usepackage{microtype}
\PassOptionsToPackage{warn}{textcomp}
\usepackage{textcomp}
\usepackage{mathptmx}
\usepackage{times}

\usepackage{cite}
\usepackage{booktabs}
\usepackage{amsmath}
\usepackage{amssymb}
\usepackage{tikz}
\usetikzlibrary{positioning, arrows.meta, fit, backgrounds}

\onlineid{1281}
\vgtccategory{Research}
\vgtcpapertype{system}

\title{mlx-vis: GPU-Native Dimensionality Reduction on Apple Silicon}

\author{Han Xiao}
\affiliation{\parbox{\columnwidth}{\centering Elastic\\ E-mail: han.xiao@elastic.co}}
\authorfooter{
\item Han Xiao is with Elastic. E-mail: han.xiao@elastic.co.
}

\shortauthortitle{Xiao: mlx-vis: GPU-Native Dimensionality Reduction on Apple Silicon}

\abstract{Dimensionality reduction is a foundational tool for visualizing high-dimensional data, yet its reference implementations span a fragmented stack of CPU-bound Python packages that leaves the Metal GPU on Apple Silicon entirely unused. We present \textbf{mlx-vis}, a library that reimplements seven widely used dimensionality reduction methods and $k$-nearest neighbor graph construction in pure MLX, with every stage---from PCA preprocessing through embedding optimization to a circle-splatting renderer---executing on GPU. On Fashion-MNIST 70K, all seven methods embed in 2.1--4.6\,s on an M3 Ultra, achieving 3--13$\times$ speedups over CPU baselines while reducing the entire dependency stack to MLX and NumPy. The same pipeline scales to ten million points on a single workstation. Code at \url{https://github.com/hanxiao/mlx-vis}.%
}


\CCScatlist{
 \CCScat{I.3.6}{Computing methodologies}{Computer graphics}{Graphics systems and interfaces};
 \CCScat{I.5.3}{Computing methodologies}{Pattern recognition}{Clustering}
}

\vgtcinsertpkg

\begin{document}

\firstsection{Introduction}
\maketitle

%% \section{Introduction}

Dimensionality reduction transforms high-dimensional data into two-dimensional representations that reveal cluster structure, continuity, and outliers. The field has produced a rich family of methods: t-SNE~\cite{vandermaaten2008tsne} and UMAP~\cite{mcinnes2018umap} preserve local neighborhoods through neighbor embedding, PaCMAP~\cite{wang2021pacmap} and TriMap~\cite{amid2019trimap} use triplet-based objectives, LocalMAP~\cite{wang2025localmap} extends PaCMAP with dynamic graph adjustment, DREAMS~\cite{kury2025dreams} hybridizes t-SNE with PCA regularization, and CNE~\cite{damrich2023cne} unifies neighbor embedding under contrastive learning.

The reference implementations are distributed across independent Python packages with heterogeneous dependencies: \texttt{umap-learn}~\cite{umap_learn} relies on \texttt{numba} and \texttt{pynndescent}, \texttt{openTSNE}~\cite{openTSNE} wraps C and Cython extensions, and \texttt{pacmap} and \texttt{trimap} each carry their own trees. All are CPU-bound, running gradient updates and neighbor searches on the CPU even when GPU hardware is available, leaving the Metal GPU and unified memory on Apple Silicon entirely untapped. GPU-accelerated alternatives exist for CUDA -- notably cuML~\cite{raschka2020cuml} -- but no equivalent targets Apple Silicon.

This paper presents \texttt{mlx-vis}, a library that addresses both the dependency fragmentation and the hardware utilization gap through a single design principle: reimplement everything in pure MLX~\cite{mlx2023}. MLX is Apple's array framework for Metal GPU, exposing a NumPy-compatible interface with lazy evaluation, JIT compilation via \texttt{@mx.compile}, and unified memory access that eliminates CPU-GPU data transfers. The resulting library depends only on MLX and NumPy.

The contribution is twofold. First, hardware-adaptive reimplementation acts as dependency compression: a single GPU-native implementation replaces the fragmented CPU stack and is both faster and lighter. Second, GPU acceleration extends beyond embedding to visualization, through a circle-splatting renderer and hardware H.264 encoder producing animations without leaving the GPU. On Fashion-MNIST 70K on an M3 Ultra, the full pipeline -- raw data to a rendered 800-frame $1000\times1000$ animation -- completes in under 6 seconds.

\begin{figure*}[tbhp]
\centering
\begin{tabular}{cccc}
\includegraphics[width=0.22\textwidth]{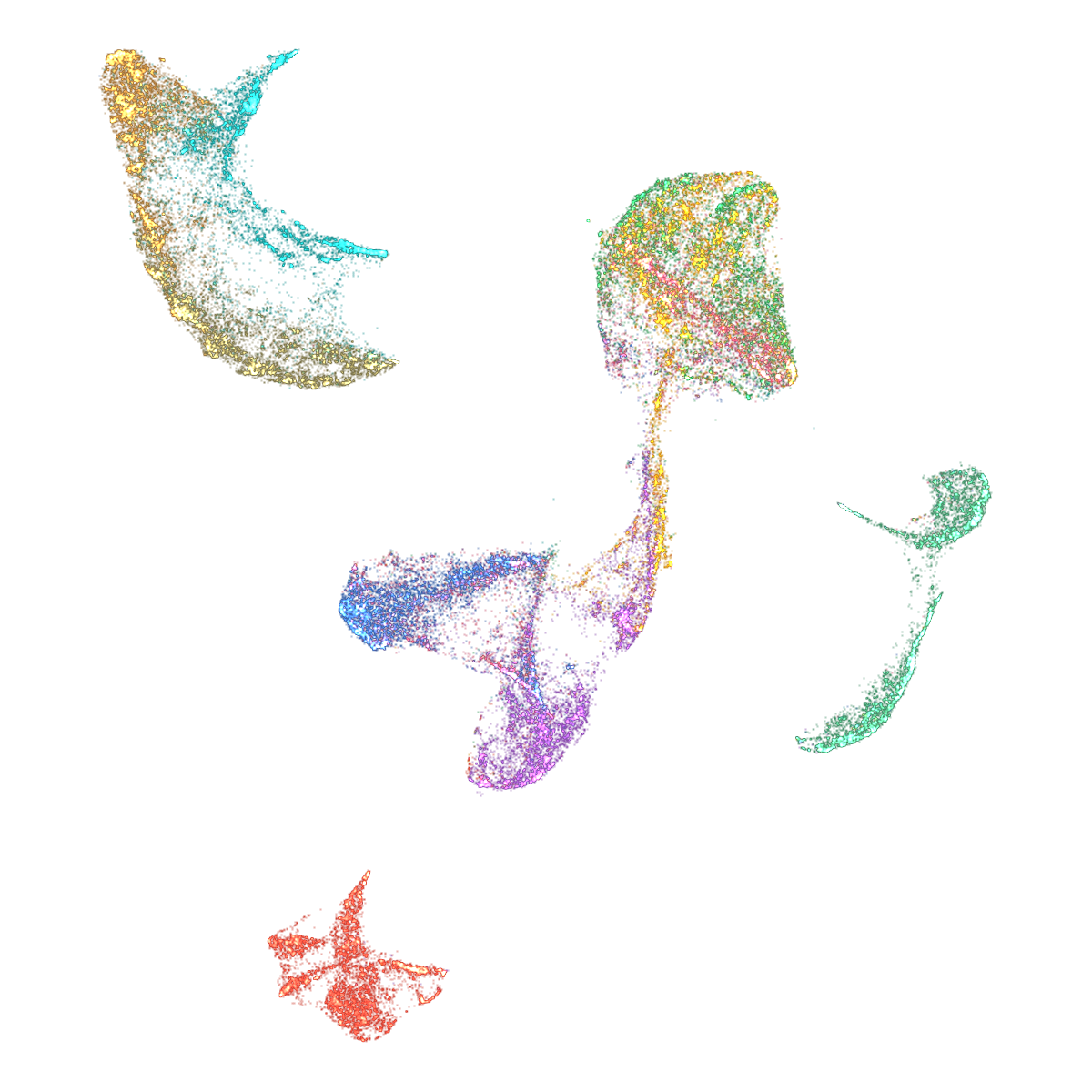} &
\includegraphics[width=0.22\textwidth]{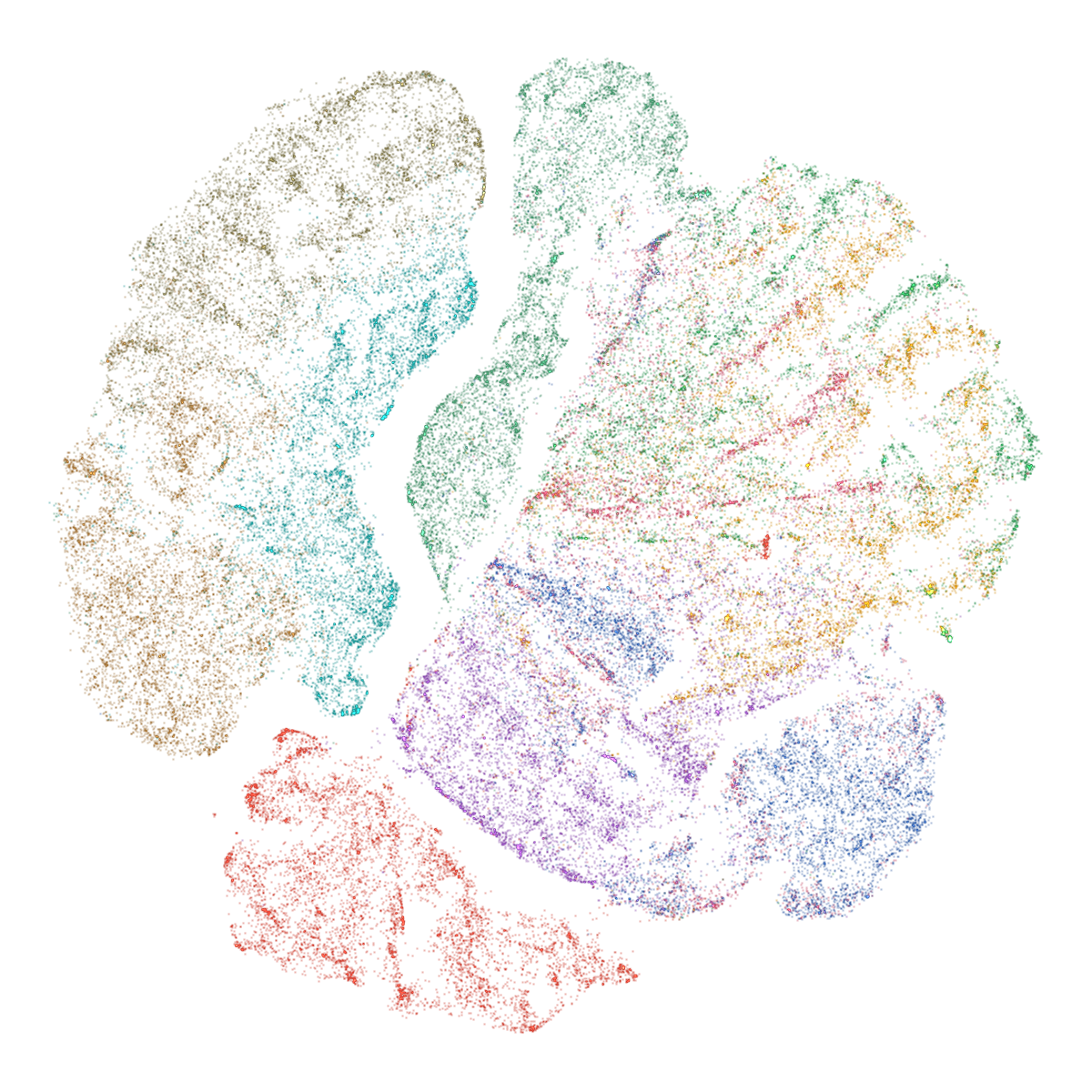} &
\includegraphics[width=0.22\textwidth]{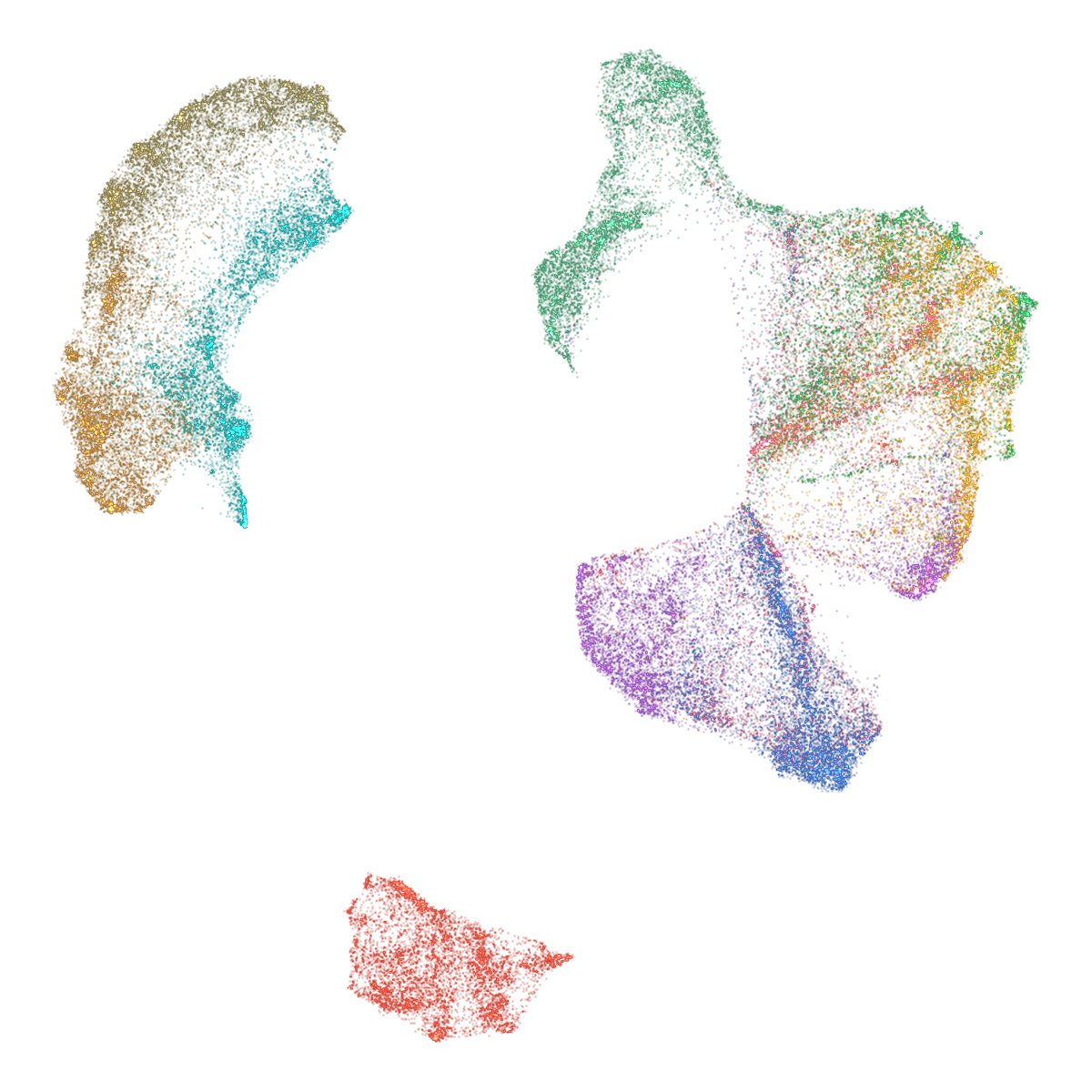} &
\includegraphics[width=0.22\textwidth]{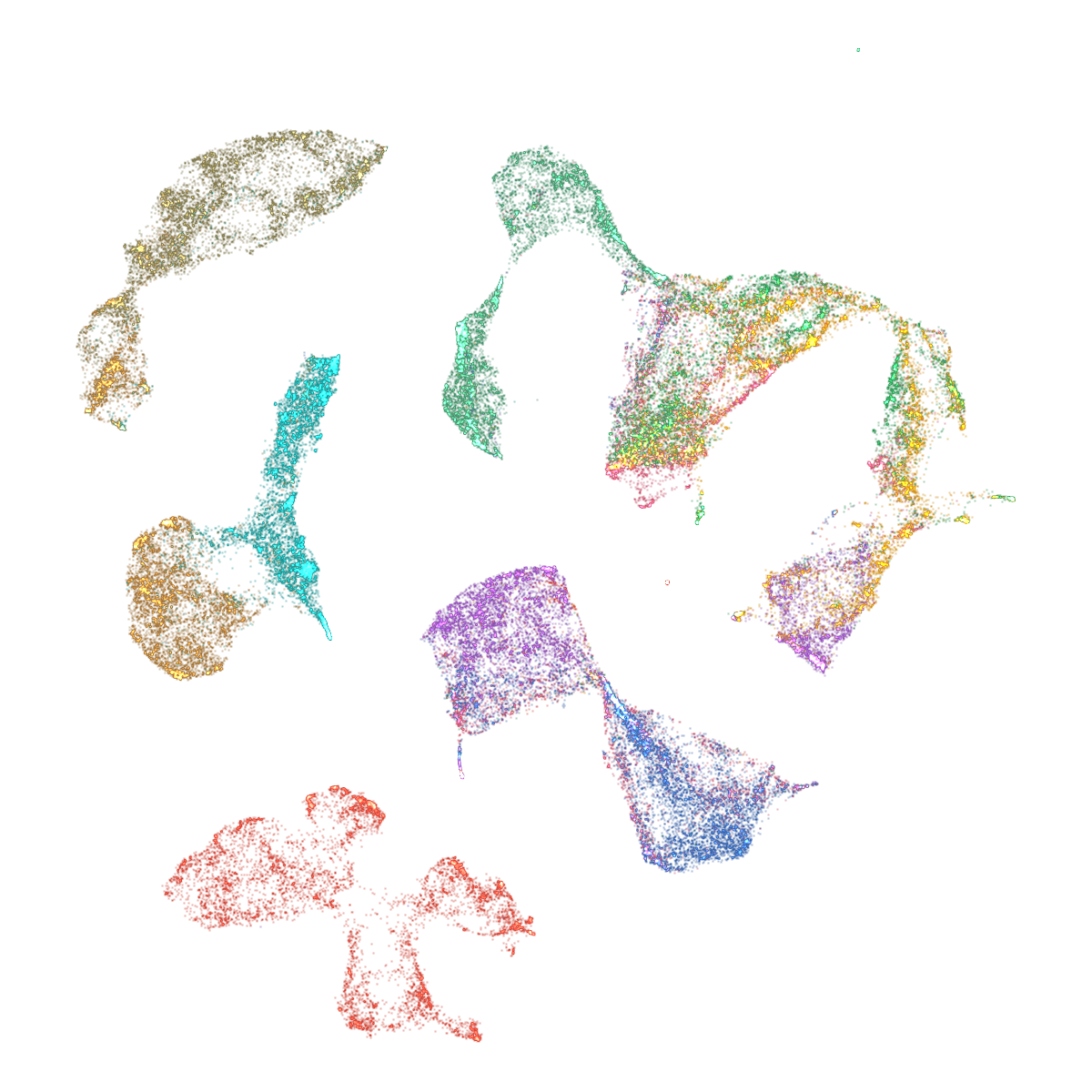} \\
UMAP & t-SNE & PaCMAP & LocalMAP \\[1em]
\includegraphics[width=0.22\textwidth]{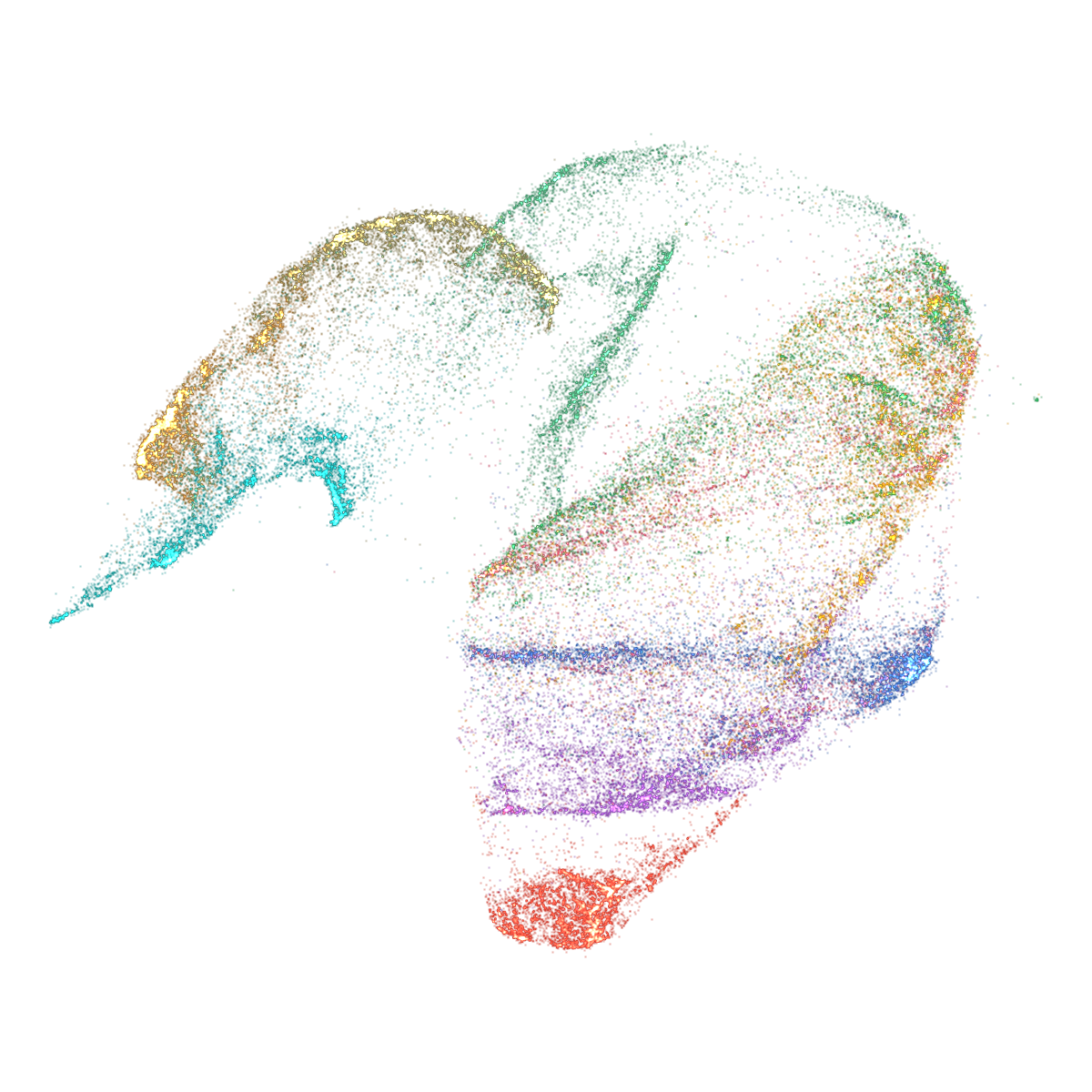} &
\includegraphics[width=0.22\textwidth]{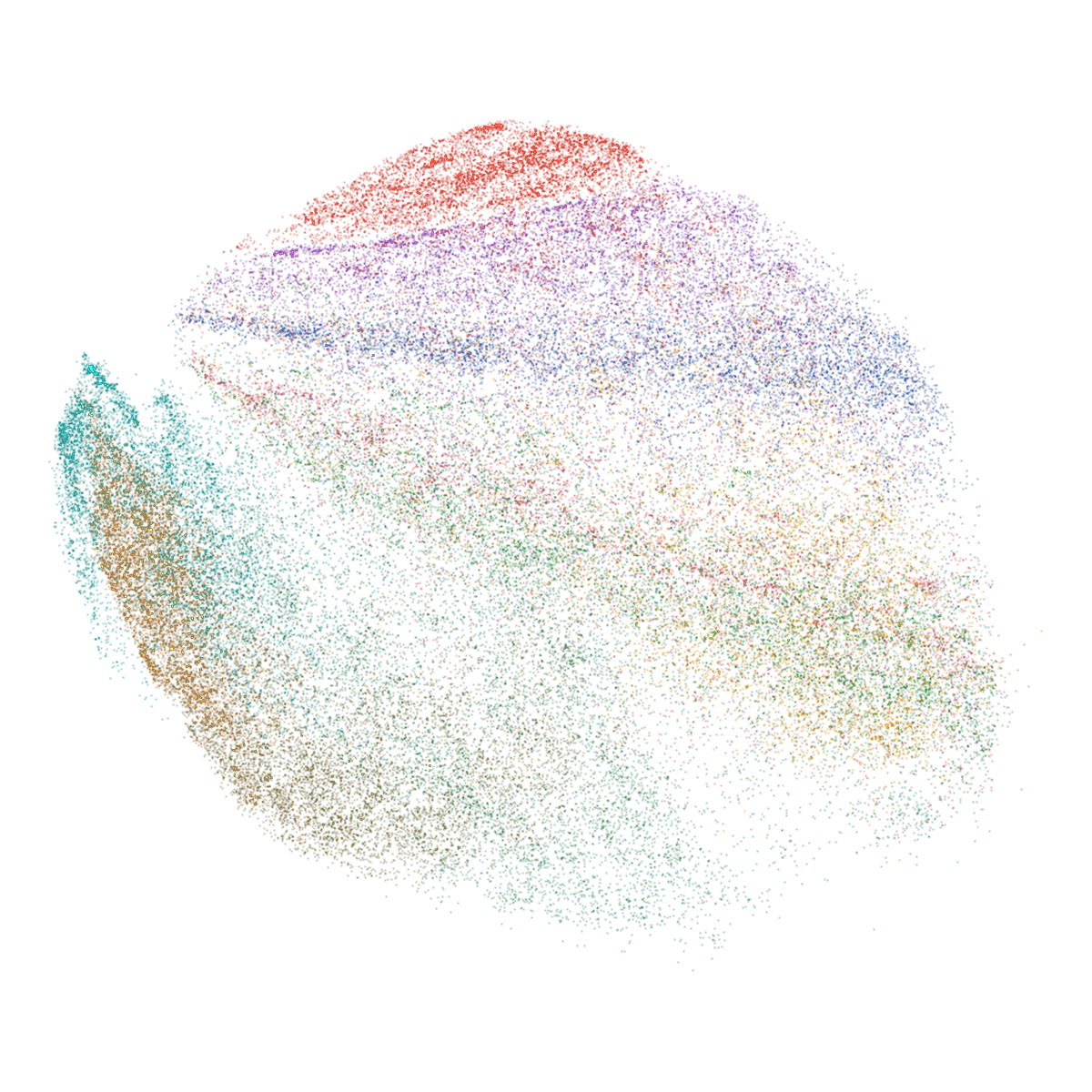} &
\includegraphics[width=0.22\textwidth]{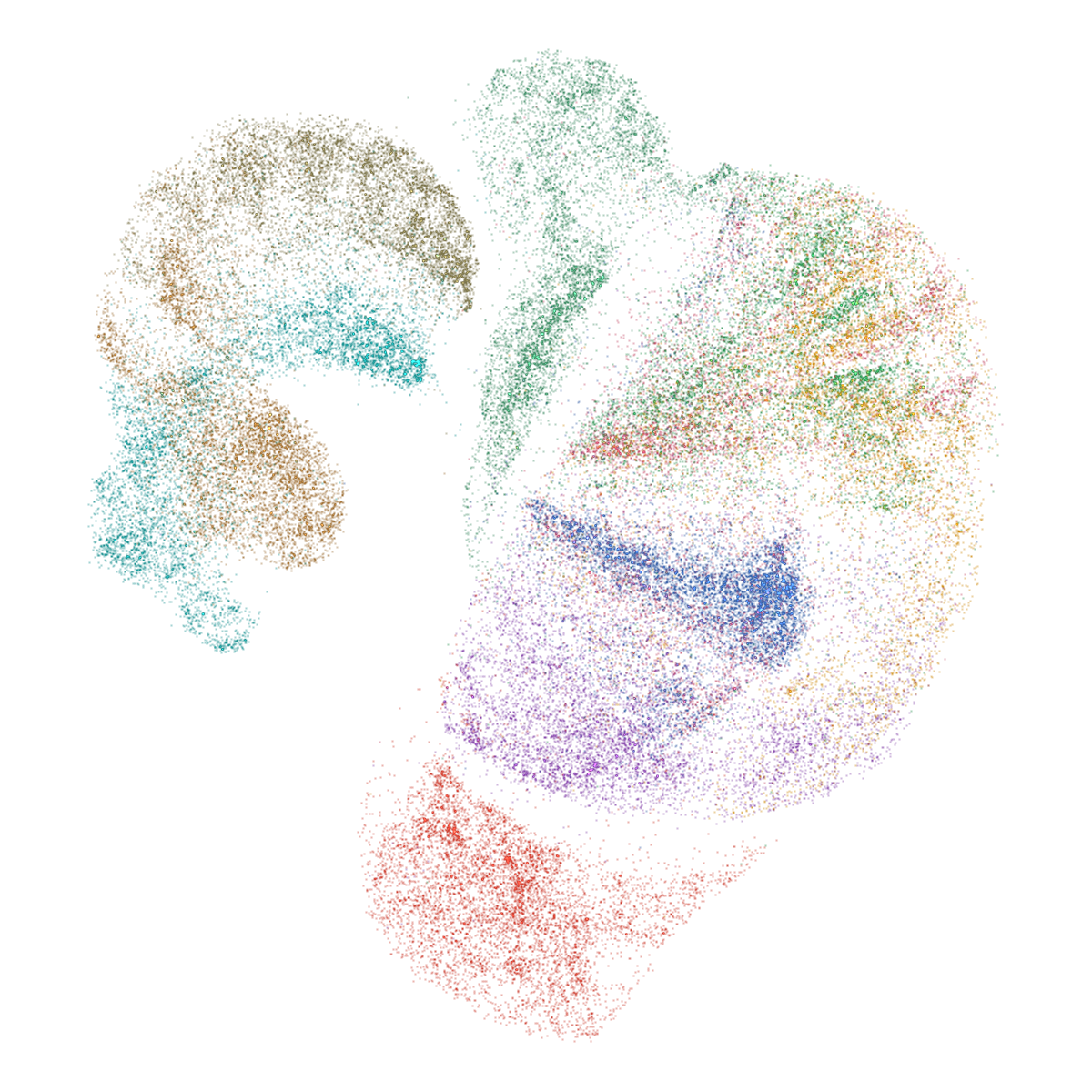} &
\includegraphics[width=0.22\textwidth]{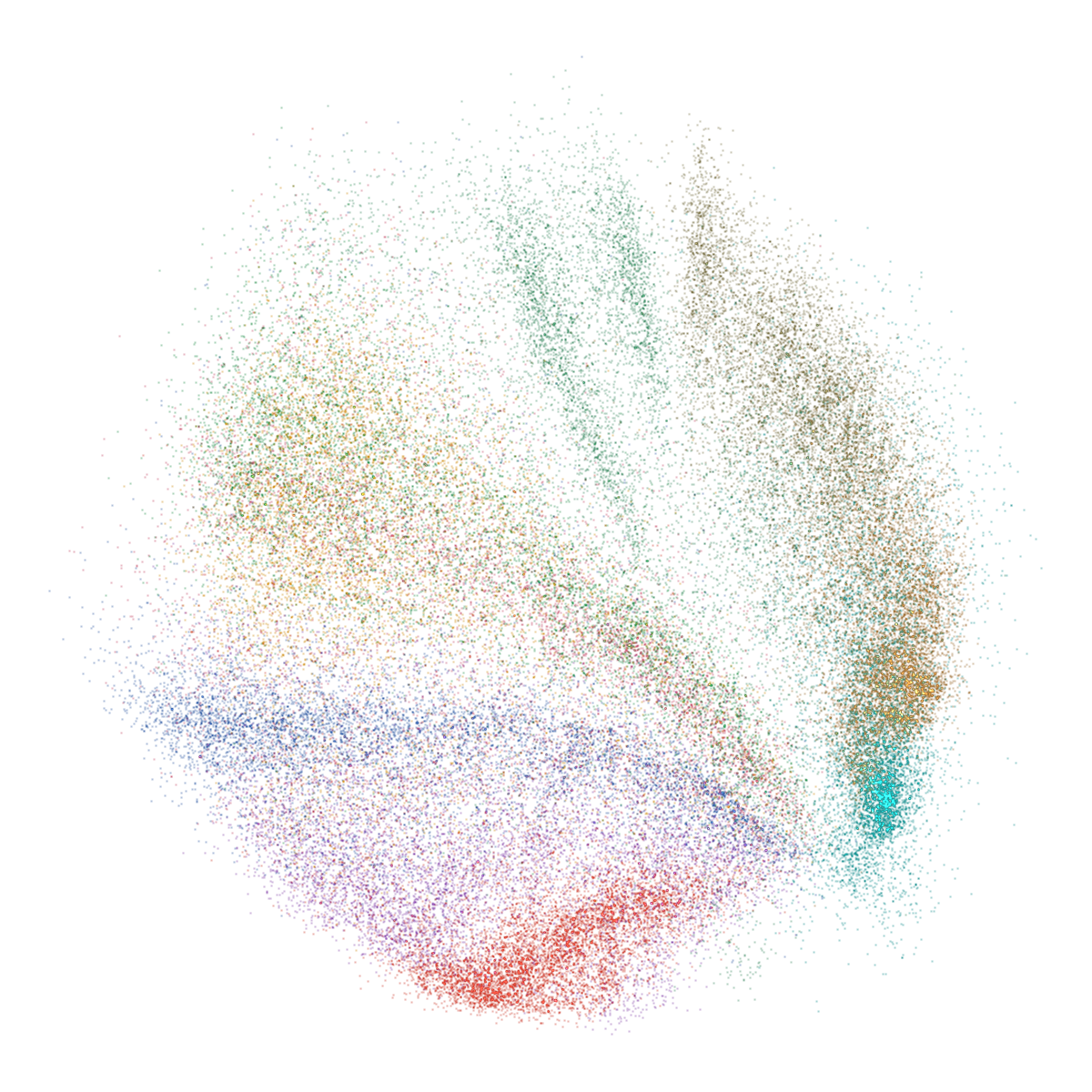} \\
TriMap & DREAMS & CNE & MMAE
\end{tabular}
\caption{Fashion-MNIST 70K embeddings produced by \texttt{mlx-vis} (MMAE shown for completeness; quantitative results in Sections~3--3.3 cover the seven validated methods).}
\label{fig:embeddings}
\end{figure*}

\section{Design and Implementation}

\subsection{Architecture}

Figure~\ref{fig:arch} shows the \texttt{mlx-vis} pipeline. Each method is a self-contained module sharing a common NNDescent implementation for $k$-NN graph construction, with GPU rendering in a separate module. The public API consists of nine algorithm classes and six visualization functions, all following the same convention: instantiate with hyperparameters, call \texttt{fit\_transform(X)} on an $n \times d$ array, and receive an $n \times 2$ embedding. An \texttt{epoch\_callback} parameter receives the current embedding each iteration, enabling animation without modifying the optimization loop.

\begin{figure}[t]
\centering
\resizebox{\columnwidth}{!}{%
\begin{tikzpicture}[
    >=Stealth,
    node distance=0.5cm,
    box/.style={draw=black!60, fill=white, rounded corners=2pt, minimum height=0.7cm, minimum width=1.8cm, font=\small\sffamily},
    sbox/.style={draw=black!50, fill=white, rounded corners=1.5pt, minimum height=0.55cm, minimum width=1.3cm, font=\scriptsize\sffamily},
    arr/.style={->, black!60, semithick},
]
\node[box] (in) {$\mathbf{X} \in \mathbb{R}^{n \times d}$};
\node[box, right=1.2cm of in] (nn) {NNDescent};
\node[sbox, right=1cm of nn, yshift=0.35cm] (u) {UMAP};
\node[sbox, right=0.1cm of u] (t) {t-SNE};
\node[sbox, right=0.1cm of t] (p) {PaCMAP};
\node[sbox, right=0.1cm of p] (lm) {LocalMAP};
\node[sbox, below=0.1cm of u] (tr) {TriMap};
\node[sbox, below=0.1cm of t] (d) {DREAMS};
\node[sbox, below=0.1cm of p] (c) {CNE};
\node[sbox, below=0.1cm of lm] (mm) {MMAE};
\begin{scope}[on background layer]
\node[draw=black!50, fill=black!3, rounded corners=3pt, fit=(u)(mm), inner sep=4pt] (mbox) {};
\end{scope}
\node[box, right=0.8cm of mbox] (out) {$\mathbf{Y} \in \mathbb{R}^{n \times 2}$};
\node[box, below=0.8cm of out] (ren) {\texttt{GPU Renderer}};
\node[box, right=1.2cm of ren] (vid) {PNG / MP4};
\begin{scope}[on background layer]
\node[draw=black!40, fill=black!8, rounded corners=4pt,
      fit=(nn)(mbox)(out)(ren),
      inner xsep=10pt, inner ysep=10pt,
      label={[font=\small\sffamily, black!60]below:Metal GPU \quad (Apple Silicon Unified Memory)}] (gpubox) {};
\end{scope}
\draw[arr] (in) -- (nn);
\draw[arr] (nn) -- (mbox);
\draw[arr] (mbox) -- (out);
\draw[arr] (out) -- (ren);
\draw[arr] (ren) -- (vid);
\end{tikzpicture}%
}
\caption{The \texttt{mlx-vis} pipeline. All stages inside the shaded region execute on Metal GPU through MLX.}
\label{fig:arch}
\end{figure}

\subsection{GPU-Native $k$-NN via NNDescent}

Approximate $k$-nearest neighbor search is the first stage of every method. We implement NNDescent~\cite{dong2011nndescent} entirely in MLX, with the entire neighbor graph kept on GPU so that no CPU--GPU transfer occurs before embedding optimization. The algorithm initializes each point with $k$ random neighbors and iteratively refines the graph by exploring neighbors-of-neighbors. Distances are computed via GPU matrix multiplication, $\|a - b\|^2 = \|a\|^2 + \|b\|^2 - 2a^{\top}b$, and top-$k$ selection uses \texttt{mx.argpartition} to avoid full sorting, with early termination when the update rate falls below $\delta = 0.015$. Reference implementations realize the same algorithm with numba-compiled Python or C extensions over CPU memory; that layer disappears here.

\subsection{Method Implementations}

All seven methods follow the standard two-phase pattern: construct a graph or sampling structure, then iteratively optimize a 2D embedding. Each implementation faithfully reproduces the published algorithm. Notable MLX-specific adaptations include the following. UMAP fits its output kernel parameters via Gauss-Newton optimization rather than scipy curve fitting. t-SNE provides an FFT-accelerated $O(n \log n)$ repulsive force variant following FIt-SNE~\cite{linderman2019fitsne}. LocalMAP's dynamic local pair resampling uses pure MLX GPU operations with \texttt{mx.argsort}-based candidate selection instead of the original per-row Python loop. CNE extracts each contrastive loss into a compiled static method for operator fusion.

\subsection{GPU Rendering Pipeline}

A distinctive aspect of \texttt{mlx-vis} is that GPU acceleration extends beyond computation to visualization. Rather than delegating to matplotlib -- which transfers data to CPU and rasterizes in software -- the library implements a circle-splatting renderer entirely in MLX.

For each point, pixel offsets within radius $R$ get linear falloff weights $w = \max(0, 1 - r/R)$; premultiplied color contributions are accumulated into a framebuffer via \texttt{mx.array.at[idx].add(vals)}, an atomic scatter-add on GPU, then a normalization pass divides color by alpha and composites over the background. For animation, hold frames reuse a single buffer, \texttt{mx.async\_eval()} overlaps rendering of frame $n{+}1$ with I/O of frame $n$, and frames are piped to \texttt{ffmpeg} with \texttt{h264\_videotoolbox} hardware encoding, keeping the path from embedding to encoded video on Apple Silicon accelerators.

\subsection{Hardware-Specific Optimizations}

Two properties of MLX and Apple Silicon are central to the performance gains.

\textbf{Lazy evaluation and operator fusion.}
MLX dispatches work to the GPU only upon \texttt{mx.eval()}; placing evaluation gates at the end of each epoch lets the framework fuse operations into fewer GPU dispatches. The \texttt{@mx.compile} decorator JIT-compiles pure functions into fused kernels, which we apply to UMAP's SGD step, t-SNE's repulsive kernel, PaCMAP's per-phase update, and CNE's per-loss gradient.

\textbf{Unified memory and zero-copy access.}
Apple Silicon's unified memory lets CPU and GPU access the same physical memory without explicit transfers. In \texttt{mlx-vis}, data is allocated once and accessed by GPU kernels throughout the pipeline---from PCA preprocessing through neighbor search, embedding optimization, and rendering---eliminating the transfer overhead that plagues heterogeneous CPU-GPU systems.

\section{Benchmarks}

All benchmarks use Fashion-MNIST~\cite{xiao2017fashionmnist}: 70,000 images of $28 \times 28$ pixels flattened to 784 dimensions, with z-score normalization per feature. The hardware is an Apple M3 Ultra with 512\,GB unified memory. All timings are mean $\pm$ standard deviation over 5 runs, with peak GPU power sampled at 100\,ms intervals via \texttt{macmon}.

Table~\ref{tab:bench} reports embedding time, peak GPU memory, and peak GPU power for each method. Iteration counts follow each method's published default (500 epochs for UMAP/t-SNE/TriMap/DREAMS/CNE; the canonical $(100,100,300)$ phase schedule for PaCMAP/LocalMAP). Compared to multi-threaded CPU baselines on the same Mac Studio (all CPU cores; package versions in the Reproducibility paragraph below), \texttt{mlx-vis} achieves speedups of 3.4$\times$ over \texttt{umap-learn}, 12.6$\times$ over \texttt{openTSNE}, 1.7$\times$ over \texttt{pacmap}, and 7.2$\times$ over \texttt{trimap}, with all hyperparameters held to each method's defaults. The GPU rendering pipeline adds $1.43 \pm 0.21$\,s for an 800-frame animation at $1000 \times 1000$; rendering the same frames with a standard \texttt{matplotlib} scatter pass takes $\sim$124\,s ($\sim$87$\times$ slower), as it rasterizes in software on CPU. We report this single point only and do not claim superiority over GPU pipelines such as datashader or WebGL tools, which we leave to future work. All seven methods complete in 2.1--4.6 seconds.

\begin{table}[t]
\caption{Embedding performance on Fashion-MNIST (70K points, 784 dimensions), M3 Ultra. Mean $\pm$ std over 5 runs.}
\label{tab:bench}
\centering
\begin{tabular}{@{}lrrr@{}}
\toprule
Method & \multicolumn{1}{c}{Time (s)} & \multicolumn{1}{c}{Mem (GB)} & \multicolumn{1}{c}{Power (W)} \\
\midrule
UMAP     & $2.60 \pm 0.03$ & 2.5 & 58 \\
t-SNE    & $4.59 \pm 0.00$ & 3.2 & 74 \\
PaCMAP   & $3.97 \pm 0.03$ & 3.0 & 70 \\
LocalMAP & $4.18 \pm 0.02$ & 3.0 & 68 \\
TriMap   & $2.11 \pm 0.03$ & 2.6 & 67 \\
DREAMS   & $4.63 \pm 0.01$ & 3.2 & 73 \\
CNE      & $2.97 \pm 0.02$ & 2.3 & 61 \\
\bottomrule
\end{tabular}
\end{table}

The resource profile is notable. At the Fashion-MNIST 70K scale, peak GPU memory ranges from 2.3 to 3.2\,GB, well within the capacity of any Apple Silicon device -- including the base M1 with 8\,GB unified memory -- and peak power stays between 58 and 74\,W. We measured power on Apple Silicon only and make no direct comparison to CUDA hardware, for which we report no measurements. The combination of low memory footprint, low power draw, and single-digit-second latency makes \texttt{mlx-vis} suitable for interactive exploration on consumer Apple Silicon at this scale; the large-scale regime of Section~3.4 has a substantially larger memory envelope, discussed there.

\textbf{Is the gain from MLX or from the GPU?}
To separate the contribution of MLX from that of Metal GPU execution in general, we reimplement the UMAP embedding optimizer in PyTorch using the Metal Performance Shaders (MPS) backend, holding the fuzzy-simplicial-set graph, spectral initialization, epoch count, negative-sample rate, and gradient expression identical, and timing only the 500-epoch optimization loop on Fashion-MNIST 70K. Table~\ref{tab:mps} shows that moving to the GPU is the dominant factor -- both MLX and PyTorch/MPS are far faster than the CPU baselines of Table~\ref{tab:bench} -- but MLX retains a measurable edge: $2.6\times$ over eager PyTorch/MPS and $1.17\times$ over \texttt{torch.compile}, attributable to MLX's lazy-evaluation kernel fusion and lower dispatch overhead on Apple Silicon.

\begin{table}[t]
\caption{MLX vs. PyTorch/MPS on the \emph{identical} UMAP optimization loop (Fashion-MNIST 70K, 500 epochs, M3 Ultra, mean $\pm$ std over 5 runs). Same graph, init, and gradient; only the array framework differs.}
\label{tab:mps}
\centering
\begin{tabular}{@{}lrr@{}}
\toprule
Backend & Time (s) & vs. MLX \\
\midrule
MLX (\texttt{@mx.compile})        & $0.42 \pm 0.01$ & $1.0\times$ \\
PyTorch/MPS (eager)               & $1.07 \pm 0.04$ & $2.6\times$ \\
PyTorch/MPS (\texttt{torch.compile}) & $0.49 \pm 0.03$ & $1.17\times$ \\
\bottomrule
\end{tabular}
\end{table}

\textbf{Reproducibility.} All \texttt{mlx-vis} timings use \texttt{mlx-vis} 0.7.0 with MLX on macOS, seeds 0--9 for Table~\ref{tab:bench} and 0--4 for Table~\ref{tab:mps}. CPU baselines run \texttt{umap-learn} 0.5.11, \texttt{openTSNE} 1.0.4, \texttt{pacmap} 0.9.1, and \texttt{trimap} 1.1.5 under Python 3.12 / NumPy 2.4 with the default Accelerate BLAS and \texttt{n\_jobs=-1} (all CPU cores). Reported embedding time is the end-to-end \texttt{fit\_transform} wall clock -- including PCA preprocessing, NNDescent graph construction, and embedding optimization -- measured after one warmup run so that \texttt{@mx.compile} JIT compilation is excluded; the optimization loop alone accounts for $0.42$\,s of UMAP's $2.60$\,s, with the remainder dominated by $k$-NN graph construction. Memory and power are sampled via \texttt{macmon} at 100\,ms intervals.

\subsection{Dependency Elimination as Resource Reduction}

The reference stack requires six compiled or JIT-compiled packages (numba, scipy, sklearn, pynndescent, Cython extensions, matplotlib) totaling hundreds of megabytes; \texttt{mlx-vis} replaces it with MLX and NumPy. Each eliminated dependency tracks back to a CPU-era assumption: numba's JIT is replaced by \texttt{@mx.compile} on GPU, scipy's curve fitting and sparse linear algebra by direct MLX expressions, and matplotlib by the GPU circle-splatting pipeline, making these dependencies structurally unnecessary rather than just practically avoidable.

\subsection{Embedding Quality}

Beyond speed, we evaluate whether the GPU-native implementations preserve embedding quality. Table~\ref{tab:quality} reports three standard metrics on Fashion-MNIST 70K for all seven methods: $k$-NN preservation (fraction of $k{=}10$ high-dimensional neighbors preserved in the 2D embedding) and trustworthiness/continuity~\cite{venna2006trustworthiness}. Trustworthiness penalizes 2D neighbors that are distant in the original space; continuity is its dual, penalizing high-dimensional neighbors that are lost in the 2D embedding. To match the standard definitions and avoid biases from approximate neighbor search, all rankings used by these metrics are computed from \emph{exact} pairwise Euclidean distances on GPU (chunked matrix multiplication on the high-dimensional input and brute-force on the 2D embedding).

All seven methods achieve trustworthiness and continuity above 0.997, indicating that the 2D neighborhoods are overwhelmingly faithful to the high-dimensional structure. Under the more discriminating $k$-NN preservation metric, t-SNE attains the highest score at 0.143, followed by LocalMAP (0.122), UMAP (0.118), TriMap (0.084), PaCMAP (0.081), CNE (0.065), and DREAMS (0.049). The ranking matches the qualitative observation that t-SNE and LocalMAP produce the tightest local clusters on Fashion-MNIST.

To verify that the MLX reimplementations match their reference counterparts rather than merely falling in a plausible range, Table~\ref{tab:qualref} reports a like-for-like comparison of trustworthiness and continuity ($k{=}15$) against the reference CPU packages on a common Fashion-MNIST 10K subset under matched hyperparameters and seed. \texttt{mlx-vis} is within $\pm0.004$ of the reference on UMAP, t-SNE, and PaCMAP for both metrics, and substantially exceeds the \texttt{trimap} reference (0.974 vs.\ 0.865 trustworthiness). This confirms that the GPU-native reimplementation does not sacrifice quality for speed.

\begin{table}[t]
\caption{Embedding quality vs.\ reference implementations on Fashion-MNIST 10K ($k{=}15$, matched hyperparameters and seed). Trust: trustworthiness; Cont: continuity. mlx (this work) vs.\ ref (CPU package).}
\label{tab:qualref}
\centering
\begin{tabular}{@{}lcccc@{}}
\toprule
 & \multicolumn{2}{c}{Trust} & \multicolumn{2}{c}{Cont} \\
\cmidrule(lr){2-3}\cmidrule(lr){4-5}
Method & mlx & ref & mlx & ref \\
\midrule
UMAP   & 0.979 & 0.979 & 0.982 & 0.986 \\
t-SNE  & 0.990 & 0.989 & 0.985 & 0.985 \\
PaCMAP & 0.973 & 0.972 & 0.984 & 0.984 \\
TriMap & \textbf{0.974} & 0.865 & \textbf{0.987} & 0.900 \\
\bottomrule
\end{tabular}
\end{table}

\begin{table}[t]
\caption{Embedding quality on Fashion-MNIST 70K ($k{=}10$). All ranks computed from \emph{exact} pairwise Euclidean distances on GPU. kNN-P: $k$-NN preservation; Trust: trustworthiness; Cont: continuity.}
\label{tab:quality}
\centering
\begin{tabular}{@{}lccc@{}}
\toprule
Method   & kNN-P & Trust & Cont \\
\midrule
UMAP     & 0.118 & 0.998 & 0.999 \\
t-SNE    & \textbf{0.143} & \textbf{0.998} & \textbf{0.999} \\
PaCMAP   & 0.082 & 0.998 & 0.998 \\
LocalMAP & 0.122 & 0.998 & 0.999 \\
TriMap   & 0.084 & 0.998 & 0.998 \\
DREAMS   & 0.049 & 0.998 & 0.998 \\
CNE      & 0.065 & 0.998 & 0.998 \\
\bottomrule
\end{tabular}
\end{table}

\subsection{Scaling Behavior}

\begin{figure}[t]
\centering
\includegraphics[width=\columnwidth]{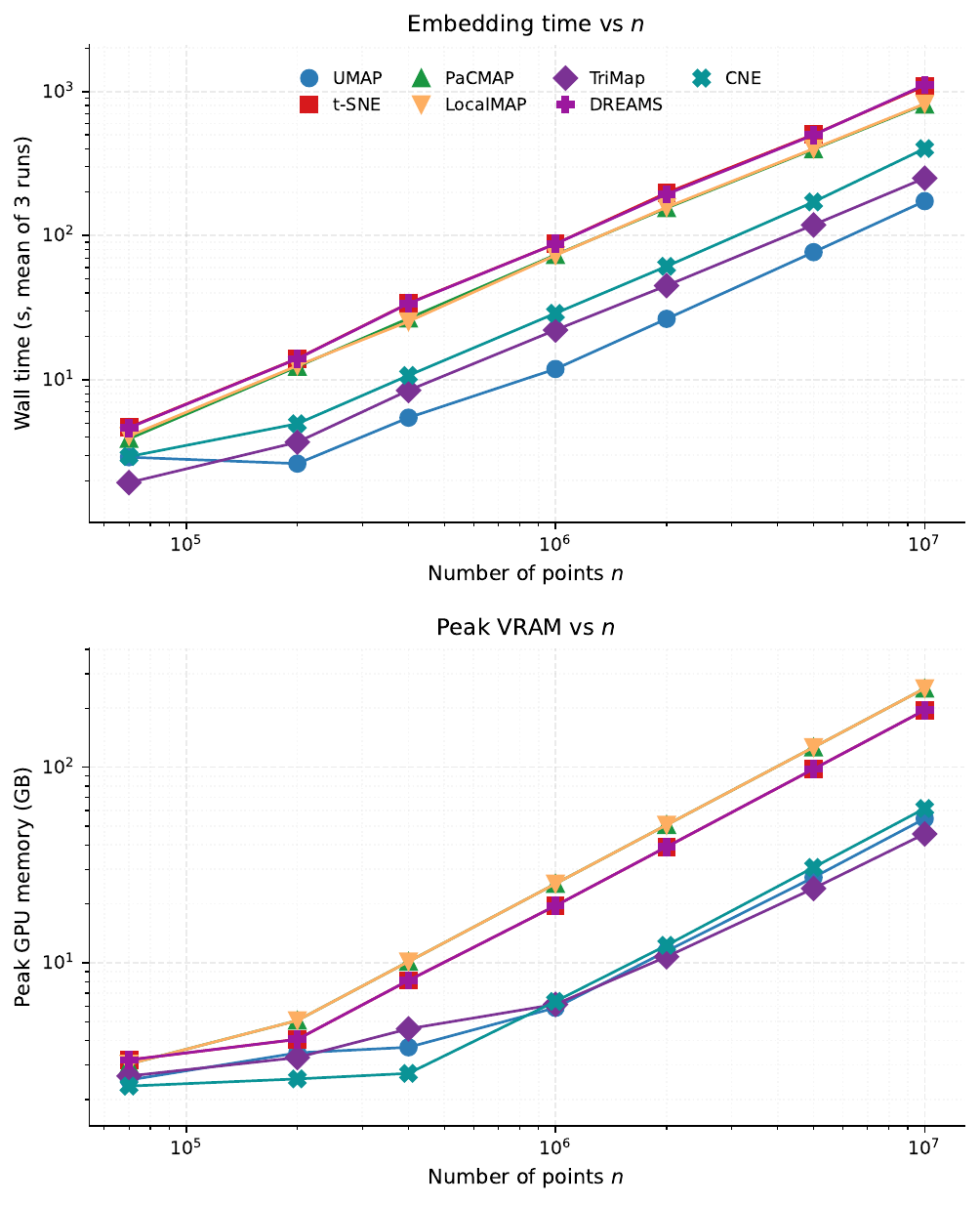}
\caption{Scaling behavior on GloVe-Twitter ($n$ from 70K to 10M points, 200-d) on a single M3 Ultra Mac Studio with 512\,GB unified memory~\cite{apple_macstudio_m3ultra}. Top: wall-clock embedding time. Bottom: peak GPU memory. Error bars (mean $\pm$ std over 3 runs) are smaller than the markers.}
\label{fig:scaling}
\end{figure}

To characterize how the GPU-native pipeline scales beyond Fashion-MNIST, we benchmark all methods on GloVe-Twitter 200-d~\cite{pennington2014glove} subsampled at seven sizes from $7\!\times\!10^4$ up to $10^7$ points. Each cell is the mean of 3 runs at default iteration counts on a single M3 Ultra. Figure~\ref{fig:scaling} reports both wall-clock time and peak GPU memory.

All seven methods complete on $10^7$ points within a single M3 Ultra session, with no out-of-memory failures across the 56 cells of the sweep. UMAP embeds 10 million points in $173.6$\,s; the slowest, DREAMS, takes $1110$\,s ($\sim$18.5 minutes). Wall-clock time scales close to log-linearly in $n$, consistent with NNDescent's $O(n^{1.14})$ empirical complexity~\cite{dong2011nndescent} dominating at large $n$ while GPU dispatch overhead amortizes at small $n$. Triplet-based methods (TriMap, PaCMAP, LocalMAP) take $\sim$2--3$\times$ less time than the FFT repulsive-force methods (t-SNE, DREAMS) at every scale. Peak GPU memory grows roughly linearly for the triplet methods (TriMap: $45.6$\,GB at $10^7$ points) and super-linearly for the repulsive-force methods (t-SNE, DREAMS: $\sim$$196$\,GB), which dominates the envelope at the largest scale. A single unified-memory workstation thus absorbs a $140\times$ growth in dataset size relative to Section~3 while running every method end-to-end on GPU, where CPU-bound implementations would require out-of-core or distributed setups.

The $\sim$196\,GB peak is, however, a property of a 512\,GB workstation, not a consumer laptop. Because MLX uses unified memory, an oversized run terminates with an out-of-memory error rather than silently spilling to disk. \texttt{mlx-vis} exposes two knobs to trade memory for time before that point: a \texttt{pca\_dim} preprocessing dimension bounding the input working set, and chunked pairwise-distance evaluation scaling the transient $k$-NN footprint. The practical ceiling on a given device thus degrades gracefully with these settings rather than failing abruptly.

\subsection{Cross-Domain Generalization}

To confirm the per-iteration cost is not domain-specific, Table~\ref{tab:crossdata} reports timing and peak GPU memory at fixed scale ($n\!\approx\!50$--$70$K) across three modalities -- greyscale digits (MNIST, $784$-d), natural images (CIFAR-10, $3072$-d), and word vectors (GloVe-Twitter, $200$-d) -- for three representative methods at the Table~\ref{tab:bench} settings. All cells complete in under 6 seconds with peak memory below 4\,GB. Input dimensionality has small but consistent impact (200-d UMAP runs $\sim$30\% faster than 784-d), and methods retain their relative ordering across all domains, indicating per-iteration cost is set by the optimizer and $k$-NN graph, not the input's domain structure.

\begin{table}[t]
\caption{Cross-domain performance at fixed scale (mean $\pm$ std over 3 runs, M3 Ultra). Domains span images (MNIST, CIFAR-10) and pretrained word vectors (GloVe-Twitter).}
\label{tab:crossdata}
\centering
\begin{tabular}{@{}llrr@{}}
\toprule
Dataset & Method & Time (s) & Mem (GB) \\
\midrule
MNIST  & UMAP   & $2.42 \pm 0.04$ & 2.4 \\
60K    & t-SNE  & $4.07 \pm 0.01$ & 3.1 \\
784-d  & PaCMAP & $3.28 \pm 0.02$ & 3.0 \\
\midrule
CIFAR-10 & UMAP   & $2.54 \pm 0.03$ & 3.0 \\
50K      & t-SNE  & $5.69 \pm 0.02$ & 3.9 \\
3072-d   & PaCMAP & $3.73 \pm 0.02$ & 3.0 \\
\midrule
GloVe-Twitter & UMAP   & $1.77 \pm 0.04$ & 2.4 \\
70K           & t-SNE  & $5.05 \pm 0.02$ & 2.9 \\
200-d         & PaCMAP & $3.92 \pm 0.04$ & 3.0 \\
\bottomrule
\end{tabular}
\end{table}

\section{Related Work}

GPU-accelerated dimensionality reduction has been explored primarily in the CUDA ecosystem: RAPIDS cuML~\cite{raschka2020cuml} provides UMAP and t-SNE on NVIDIA GPUs with substantial speedups, but targets discrete GPUs with dedicated VRAM and requires the CUDA toolkit, making it inaccessible on Apple Silicon. The MLX ecosystem~\cite{mlx2023} has produced libraries for language models, diffusion, and speech, but to our knowledge none target dimensionality reduction. \texttt{mlx-vis} fills this gap, applying the same hardware-adaptive approach that powers on-device LLM inference to classical ML workloads.

The broader theme---that reimplementing algorithms for specific hardware can simultaneously improve performance and reduce complexity---connects to the visualization community's interest in hardware-accelerated rendering. While GPU acceleration in visualization has historically focused on rasterization and raytracing, we show the same principle applies upstream to the embedding computation, bringing sub-second interactive DR within reach on consumer Apple Silicon.

\subsection{Neural Engine Applicability}

Apple Silicon chips also include a Neural Engine (ANE), a fixed-function accelerator for low-power inference, recently programmable outside CoreML~\cite{maderix2026ane, anegpt2026}. We target the Metal GPU rather than the ANE because of three architectural mismatches: the dominant operations here are scatter-add updates on randomly sampled edges and pointer-chasing neighbor lookups, neither of which the ANE expresses; the $n \times 2$ embedding output uses less than 1\% of the ANE's 256--1024-channel SRAM tiling; and the $\sim$0.5\,ms per-call IOSurface dispatch overhead~\cite{maderix2026ane} is comparable to an entire UMAP SGD step. Metal GPU avoids all three: native scatter-add, no channel-width constraint, and zero-copy unified-memory access without kernel transitions.

\section{Conclusion}

\texttt{mlx-vis} is a systems and tools contribution: rather than a new DR algorithm or visual encoding, it shows that hardware-adaptive reimplementation is a viable strategy for resource-efficient visualization infrastructure. Through pure MLX on Apple Silicon's Metal GPU and unified memory, it delivers 3--13$\times$ speedups over CPU baselines, eliminates six heavyweight dependencies, matches reference implementations on quality metrics, and extends GPU acceleration through to rendering. The architecture is built for extension: each method shares the common NNDescent graph and GPU renderer under one \texttt{fit\_transform} / \texttt{epoch\_callback} contract, so a new method is one optimizer against existing primitives. Available via \texttt{pip}.

\bibliographystyle{abbrv-doi-hyperref-narrow}
\bibliography{refs}

\end{document}